\documentclass{article}

\usepackage[final]{neurips_2026_vericode}
\usepackage[utf8]{inputenc}   
\usepackage[T1]{fontenc}      

\usepackage{amsmath}
\usepackage{amssymb}          
\usepackage{mathtools}
\usepackage{centernot}
\usepackage{nicefrac}
\usepackage{booktabs}         
\usepackage{tabularx}
\usepackage{multirow}
\usepackage{array}
\usepackage{graphicx}
\usepackage{xcolor}
\usepackage{xspace}
\usepackage{microtype}
\usepackage{enumitem}
\usepackage{pifont}
\usepackage{soul}
\usepackage{listings}
\usepackage[most]{tcolorbox}
\usepackage{tikz}
\usetikzlibrary{arrows.meta,backgrounds,calc,fit,positioning,shapes.geometric}
\usepackage{hyperref}         
\usepackage{url}              

\newcommand{\method}{\textsc{FEAgent}\xspace}
\newcommand{\sourceprog}{P_{\mathrm{s}}}
\newcommand{\targetprog}{P_{\mathrm{t}}}
\newcommand{\sourcegraph}{G_{\mathrm{s}}}
\newcommand{\targetgraph}{G_{\mathrm{t}}}
\newcommand{\obs}{\mathcal{O}}
\newcommand{\domain}{\mathcal{D}}

\newcommand{\tests}{\mathcal{X}}
\newcommand{\sem}[1]{\left[\!\left[#1\right]\!\right]}
\newcommand{\vequiv}{\textsc{Equivalent}\xspace}
\newcommand{\vinequiv}{\textsc{Inequivalent}\xspace}
\newcommand{\vabstain}{\textsc{Unclear}\xspace}
\newcommand{\regression}{\textsc{Regression}\xspace}
\newcommand{\extension}{\textsc{Extension}\xspace}

\newcolumntype{Y}{>{\centering\arraybackslash}X}
\newcolumntype{L}[1]{>{\raggedright\arraybackslash}p{#1}}
\newcolumntype{C}[1]{>{\centering\arraybackslash}p{#1}}

\definecolor{sourceorange}{HTML}{D97706}
\definecolor{targetviolet}{HTML}{6D28D9}
\definecolor{graphteal}{HTML}{0F766E}
\definecolor{decisionblue}{HTML}{1D4ED8}
\definecolor{warningred}{HTML}{B91C1C}
\definecolor{softgray}{HTML}{F3F4F6}

\newcommand{\paragraphhead}[1]{\smallskip\noindent\textbf{#1}}

\newcommand{\EquiBenchECD}{18.0\%\xspace}   
\newcommand{\SWEBenchECD}{28.4\%\xspace}    

\title{Functionally Equivalent or Not?\\
Graph-Grounded Differential Surrogate Execution for Code Equivalence}

\workshoptitle{NeurIPS 2026 Workshop on AI for Verifiable Coding}

\author{%
   Amit Kachroo \\
 AWS AI Labs \\
   Santa Clara, California, USA \\
\texttt{amkachro@amazon.com} \\
\And
   Like Hui \\
 AWS AI Labs \\
   Santa Clara, California, USA \\
\texttt{likehui@amazon.com} \\
\And
   Haitao Mao\\
 AWS AI Labs \\
   Santa Clara, California, USA \\
\texttt{maoht@amazon.com} \\
\And
   Yuhao Zhang\\
 AWS AI Labs \\
 New York, USA \\
\texttt{yhzhangh@amazon.com} \\
\And
 Nguyen Vo\\
 AWS AI Labs \\
  Santa Clara, California, USA \\
\texttt{benvox@amazon.com} \\
}

\begin{document}

\maketitle

\begin{abstract}
Determining whether two programs  are functionally equivalent or not is central not only  to code modernization, patch validation, and refactoring, but also to code-generation evaluation in general. Yet the usual signals are incomplete: tests cover only finite inputs, textual similarity confuses implementation with behavior, and unconstrained language model judgments are difficult to audit.  Moreover, direct execution is often impossible when a program depends on an obsolete, licensed, unavailable, or unsafe environment. We introduce \method, a selective equivalence assessor agent that combines typed program-graph evidence with differential surrogate execution.  \method first aligns public interfaces and behaviorally relevant graph anchors, then issues bounded queries over call-flow, control-flow, data-flow, type, import, and effect relations. With this information in context, a branch-aware generator agent proposes discriminating inputs, and two blinded language-model surrogates independently reason about and predict source and target observables.  Every claim and predicted divergence is recorded in an evidence ledger.  A deterministic reconciler then returns \vequiv, \vinequiv, or \vabstain rather than forcing a verdict when paths are uncovered or evidence conflicts.
We evaluate \method on function-level equivalence and repository-level bug patches, where the existing oracle is a benchmark label or a passing test suite. Every disagreement with that oracle is adjudicated by direct execution, revealing errors in benchmark labels and behavioral divergences missed by unit-test-only scoring. On EquiBench, execution confirms \method's disagreements with published labels on 216 of 1{,}200 evaluated pairs (18.0\%); on SWE-bench Verified, 94 of 331 test-passing agent patches (\SWEBenchECD) diverge from the reference patch. \method thus serves as an audit layer between testing and formal verification, exposing behavioral differences that existing oracles miss while keeping its evidence reviewable and its uncertainty explicit without claiming a universal proof of equivalence.


\end{abstract}

\section{Introduction}

Software engineering repeatedly asks a deceptively simple question: did a change preserve behavior? The question appears when a patch is compared with a reference fix, a component is ported to a new language, a legacy system is modernized, or generated code is evaluated. In each case, syntactically different programs may be interchangeable, while nearly identical programs may diverge on a single boundary condition. We therefore care about \emph{observable behavior}, not textual identity.

The dominant practical oracle is a test suite. Tests are indispensable, but they establish agreement only on the executed inputs and observations. A patch can pass by exploiting an uncovered shortcut; a correct alternative can fail because a test encodes an implementation detail. Symbolic and concolic execution broaden path coverage \citep{godefroid2005dart,cadar2008klee}, and translation validation can prove strong properties for supported transformations \citep{pnueli1998translation,lopes2021alive2}. Their applicability, however, depends on executable environments, faithful models of libraries and side effects, and tractable path conditions. Those assumptions often break for incomplete repositories, cross-language migrations, legacy runtimes, proprietary dependencies, or unsafe code.

In these settings, large language models (LLMs) offer a complementary capability: they can reason about code that cannot be run locally and predict execution behavior on concrete inputs \citep{gu2024cruxeval,jin2024codemind}. But asking a model ``are these programs equivalent?'' creates a weak oracle. The model may focus on surface similarity, overlook a distant callee, silently assume a library contract, or rationalize the two programs jointly. A confident paragraph does not reveal which paths were checked, which dependency facts support a claim, or what should happen when the evidence is incomplete.

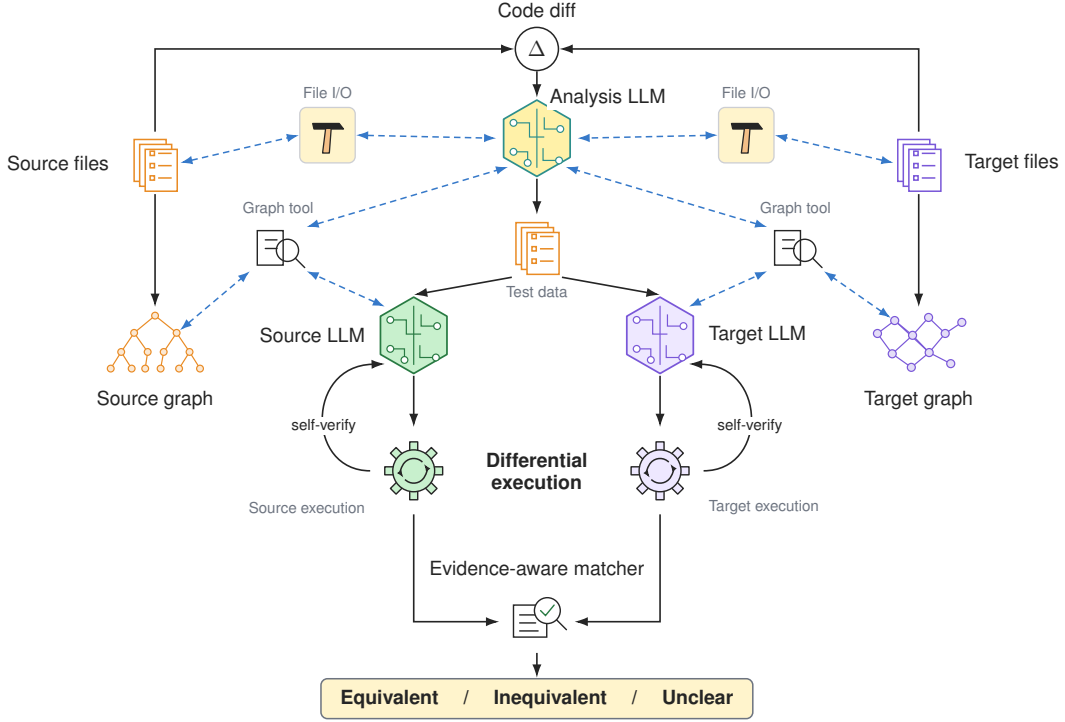
\begin{figure*}[t]
\centering
\begingroup

\definecolor{source}{HTML}{E98B24}
\definecolor{sourcefill}{HTML}{FFF0DC}
\definecolor{target}{HTML}{7957D8}
\definecolor{targetfill}{HTML}{EEE8FF}
\definecolor{analysis}{HTML}{2D8F8C}
\definecolor{analysisfill}{HTML}{FFF1A8}
\definecolor{execute}{HTML}{287A42}
\definecolor{executefill}{HTML}{C8F0CD}
\definecolor{toolblue}{HTML}{3679C9}
\definecolor{ink}{HTML}{252525}
\definecolor{muted}{HTML}{6D7480}
\definecolor{paper}{HTML}{FFFFFF}
\definecolor{decisionfill}{HTML}{FFF4CC}

\tikzset{
  every node/.style={
    font=\sffamily,
    text=ink,
    align=center
  },
  flow/.style={
    -{Latex[length=2.2mm,width=1.5mm]},
    draw=ink,
    line width=0.75pt
  },
  tool interaction/.style={
    arrows={Latex[length=2mm,width=1.4mm]-
      Latex[length=2mm,width=1.4mm]},
    draw=toolblue,
    line width=0.7pt,
    dashed,
    dash pattern=on 3pt off 2pt
  },
  panel/.style={
    draw=muted!35,
    fill=paper,
    rounded corners=3pt,
    line width=0.6pt
  },
  result/.style={
    draw=muted,
    fill=decisionfill,
    rounded corners=3pt,
    line width=0.7pt,
    inner xsep=9pt,
    inner ysep=5pt,
    font=\sffamily\bfseries\small
  },
  caption text/.style={
    font=\sffamily\small,
    align=center,
    fill=paper,
    inner sep=1pt
  },
  annotation text/.style={
    font=\sffamily\scriptsize,
    text=muted,
    fill=paper,
    inner sep=1pt
  }
}

\newcommand{\FileIcon}[3][source]{%
  \node[minimum width=8mm,minimum height=10mm,inner sep=0pt] (#2) at #3 {};
  \begin{scope}[shift={(#2.center)},draw=#1,line width=0.7pt]
    \draw[fill=paper] (-0.34,-0.30) rectangle (0.18,0.38);
    \draw[fill=paper] (-0.25,-0.38) rectangle (0.27,0.30);
    \draw[fill=paper] (-0.16,-0.46) rectangle (0.36,0.22);
    \foreach \y in {-0.28,-0.08,0.12} {
      \draw (-0.08,\y) rectangle +(0.07,0.07);
      \draw (0.05,\y+0.035) -- (0.27,\y+0.035);
    }
  \end{scope}
}

\newcommand{\LLMIcon}[4]{%
  \node[
    circle,
    minimum size=13mm,
    draw=none,
    fill=none,
    inner sep=0pt
  ] (#2) at #1 {};
  \begin{scope}[shift={(#2.center)},draw=#3,line width=0.55pt]
    \path[
      draw=#3,
      fill=#4,
      line width=0.8pt,
      line join=round
    ]
      (0,0.62)
      -- (0.54,0.31)
      -- (0.54,-0.31)
      -- (0,-0.62)
      -- (-0.54,-0.31)
      -- (-0.54,0.31)
      -- cycle;
    \draw (0,-0.48) -- (0,0.48);
    \draw (-0.38,0.25) -- (-0.12,0.25) -- (-0.12,0.02) -- (0.12,0.02);
    \draw (0.12,0.34) -- (0.12,0.15) -- (0.38,0.15);
    \draw (-0.38,-0.22) -- (-0.15,-0.22) -- (-0.15,-0.40);
    \draw (0.10,-0.16) -- (0.10,-0.36) -- (0.36,-0.36);
    \foreach \x/\y in {
      -0.38/0.25,
      0.38/0.15,
      -0.38/-0.22,
      -0.15/-0.40,
      0.36/-0.36
    } {
      \fill[paper] (\x,\y) circle (0.055);
      \draw (\x,\y) circle (0.055);
    }
  \end{scope}
}

\newcommand{\FileToolIcon}[2]{%
  \node[
    panel,
    fill=decisionfill,
    minimum width=9mm,
    minimum height=9mm,
    inner sep=0pt
  ] (#1) at #2 {};
  \begin{scope}[shift={(#1.center)},line cap=round,line join=round]
    \draw[draw=ink,fill=ink,line width=0.6pt]
      (-0.24,0.19) -- (0.19,0.19) -- (0.26,0.10) --
      (-0.22,0.10) -- cycle;
    \draw[draw=ink,fill=source!55,line width=0.6pt]
      (-0.03,0.11) -- (0.06,0.11) -- (0.03,-0.29) --
      (-0.08,-0.29) -- cycle;
  \end{scope}
}

\newcommand{\GraphToolIcon}[2]{%
  \node[minimum width=9mm,minimum height=7mm,inner sep=0pt] (#1) at #2 {};
  \begin{scope}[shift={(#1.center)},draw=ink,line width=0.65pt,line cap=round]
    \draw[fill=paper] (-0.34,-0.26) rectangle (0.08,0.30);
    \draw (-0.21,0.15) -- (-0.02,0.15);
    \draw (-0.21,0.01) -- (-0.02,0.01);
    \draw (0.15,-0.02) circle (0.18);
    \draw (0.28,-0.15) -- (0.43,-0.30);
  \end{scope}
}

\newcommand{\GraphIcon}[4]{%
  \node[minimum width=16mm,minimum height=11mm,inner sep=0pt] (#2) at #3 {};
  \begin{scope}[shift={(#2.center)},draw=#1,line width=0.55pt]
    \ifx#4t
      \draw (0,0.42) -- (-0.34,0.14) -- (-0.55,-0.17);
      \draw (-0.34,0.14) -- (-0.12,-0.17);
      \draw (0,0.42) -- (0.35,0.14) -- (0.14,-0.17);
      \draw (0.35,0.14) -- (0.56,-0.17);
      \draw (-0.55,-0.17) -- (-0.72,-0.44);
      \draw (-0.55,-0.17) -- (-0.38,-0.44);
      \draw (-0.12,-0.17) -- (-0.16,-0.44);
      \draw (0.14,-0.17) -- (0.09,-0.44);
      \draw (0.56,-0.17) -- (0.39,-0.44);
      \draw (0.56,-0.17) -- (0.73,-0.44);
      \foreach \x/\y in {
        0/0.42,
        -0.34/0.14,
        0.35/0.14,
        -0.55/-0.17,
        -0.12/-0.17,
        0.14/-0.17,
        0.56/-0.17,
        -0.72/-0.44,
        -0.38/-0.44,
        -0.16/-0.44,
        0.09/-0.44,
        0.39/-0.44,
        0.73/-0.44
      } {
        \fill[paper] (\x,\y) circle (0.055);
        \draw[fill=#1!20] (\x,\y) circle (0.055);
      }
    \else
      \draw
        (-0.67,0.13) -- (-0.39,0.34) -- (-0.10,0.10) --
        (0.16,0.39) -- (0.43,0.13) -- (0.70,0.31);
      \draw
        (-0.67,0.13) -- (-0.43,-0.20) -- (-0.10,0.10) --
        (0.17,-0.18) -- (0.43,0.13);
      \draw
        (-0.43,-0.20) -- (-0.12,-0.42) -- (0.17,-0.18) --
        (0.52,-0.36);
      \draw (-0.39,0.34) -- (0.17,-0.18);
      \foreach \x/\y in {
        -0.67/0.13,
        -0.39/0.34,
        -0.43/-0.20,
        -0.12/-0.42,
        -0.10/0.10,
        0.16/0.39,
        0.17/-0.18,
        0.43/0.13,
        0.70/0.31,
        0.52/-0.36
      } {
        \draw[fill=#1!20] (\x,\y) circle (0.065);
      }
    \fi
  \end{scope}
}

\newcommand{\ExecutionIcon}[3]{%
  \node[
    minimum width=14mm,
    minimum height=14mm,
    inner sep=0pt
  ] (#1) at #2 {};
  \begin{scope}[shift={(#1.center)},draw=ink,line width=0.65pt,
                line join=round]
    \foreach \a in {0,45,...,315} {
      \draw[
        fill=#3,
        rotate=\a
      ] (-0.065,0.31) rectangle (0.065,0.47);
    }
    \draw[fill=#3] (0,0) circle (0.34);
    \draw[
      -{Latex[length=1.25mm,width=0.9mm]},
      line width=0.7pt
    ] (-135:0.20) arc[start angle=-135,end angle=35,radius=0.20];
    \draw[
      -{Latex[length=1.25mm,width=0.9mm]},
      line width=0.7pt
    ] (45:0.20) arc[start angle=45,end angle=215,radius=0.20];
  \end{scope}
}

\newcommand{\MatcherIcon}[2]{%
  \node[minimum width=12mm,minimum height=10mm,inner sep=0pt] (#1) at #2 {};
  \begin{scope}[shift={(#1.center)},draw=ink,line width=0.65pt,line cap=round]
    \draw[fill=paper] (-0.38,-0.25) rectangle (0.16,0.27);
    \draw (-0.28,0.11) -- (0.04,0.11);
    \draw (-0.28,-0.02) -- (-0.02,-0.02);
    \draw (-0.28,-0.15) -- (0.02,-0.15);
    \draw[fill=paper] (0.17,0.18) circle (0.21);
    \draw (0.32,0.03) -- (0.48,-0.13);
    \draw[draw=execute] (0.06,0.18) -- (0.14,0.10) -- (0.28,0.27);
  \end{scope}
}

\resizebox{\textwidth}{!}{%
\begin{tikzpicture}[x=1cm,y=1cm]

  \FileIcon[source]{source-files}{(-6.2,3.15)}
  \node[caption text,left=3mm of source-files] {Source files};

  \FileIcon[target]{target-files}{(6.2,3.15)}
  \node[caption text,right=3mm of target-files] {Target files};

  \node[
    circle,
    draw=ink,
    fill=paper,
    minimum size=7mm,
    inner sep=0pt,
    line width=0.7pt
  ] (code-diff) at (0,5.0) {\(\Delta\)};
  \node[caption text,above=1mm of code-diff] {Code diff};

  \LLMIcon{(0,3.55)}{analysis-llm}{analysis}{analysisfill}
  \node[caption text] at (1.15,4.20) {Analysis LLM};

  \draw[flow] (source-files.north) |- (code-diff.west);
  \draw[flow] (target-files.north) |- (code-diff.east);
  \draw[flow] (code-diff) -- (analysis-llm);

  \FileToolIcon{source-file-tool}{(-3.4,3.6)}
  \node[annotation text,above=1mm of source-file-tool] {File I/O};
  \FileToolIcon{target-file-tool}{(3.4,3.6)}
  \node[annotation text,above=1mm of target-file-tool] {File I/O};

  \GraphToolIcon{source-graph-tool}{(-4.2,1.75)}
  \node[annotation text] at (-4.2,2.40) {Graph tool};
  \GraphToolIcon{target-graph-tool}{(4.2,1.75)}
  \node[annotation text] at (4.2,2.40) {Graph tool};

  \draw[tool interaction]
    (source-files.east) -- (source-file-tool.west);
  \draw[tool interaction]
    (source-file-tool.east) -- (analysis-llm.west);
  \draw[tool interaction]
    (target-files.west) -- (target-file-tool.east);
  \draw[tool interaction]
    (target-file-tool.west) -- (analysis-llm.east);

  \GraphIcon{source}{source-graph}{(-6.2,0.25)}{t}
  \node[caption text,below=2mm of source-graph] {Source graph};

  \GraphIcon{target}{target-graph}{(6.2,0.25)}{n}
  \node[caption text,below=2mm of target-graph] {Target graph};

  \LLMIcon{(-2.0,0.35)}{source-llm}{execute}{executefill}

  \LLMIcon{(2.0,0.35)}{target-llm}{target}{targetfill}

  \FileIcon[source]{test-data}{(0,1.80)}
  \node[annotation text,below=1mm of test-data] {Test data};

  \draw[flow]
    (source-files.south) -- (source-graph.north);
  \draw[flow]
    (target-files.south) -- (target-graph.north);

  \draw[tool interaction]
    (-5.80,0.43) -- (source-graph-tool.south west);
  \draw[tool interaction]
    (source-llm.north west) -- (source-graph-tool.south east);
  \draw[tool interaction]
    (source-graph-tool.north east) -- (analysis-llm.south west);

  \draw[tool interaction]
    (5.75,0.63) -- (target-graph-tool.south east);
  \draw[tool interaction]
    (target-llm.north east) -- (target-graph-tool.south west);
  \draw[tool interaction]
    (target-graph-tool.north west) -- (analysis-llm.south east);

  \draw[flow]
    (analysis-llm.south) -- (test-data.north);
  \draw[flow]
    (test-data.south west) -- (source-llm.north);
  \draw[flow]
    (test-data.south east) -- (target-llm.north);

  \ExecutionIcon{source-execution}{(-2.0,-1.85)}{executefill}
  \ExecutionIcon{target-execution}{(2.0,-1.85)}{targetfill}
  \node[annotation text,anchor=east] at (-2.75,-2.45)
    {Source execution};
  \node[annotation text,anchor=west] at (2.75,-2.45)
    {Target execution};

  \draw[flow]
    (source-llm.south) -- (source-execution.north);
  \draw[flow]
    (target-llm.south) -- (target-execution.north);
  \draw[flow]
    (source-execution.west)
    .. controls (-3.75,-1.85) and (-3.75,-0.15) ..
    node[
      pos=0.45,
      annotation text,
      text=ink,
      fill=paper,
      inner xsep=1.5pt,
      inner ysep=0.8pt
    ] {self-verify}
    (source-llm.south west);
  \draw[flow]
    (target-execution.east)
    .. controls (3.75,-1.85) and (3.75,-0.15) ..
    node[
      pos=0.45,
      annotation text,
      text=ink,
      fill=paper,
      inner xsep=1.5pt,
      inner ysep=0.8pt
    ] {self-verify}
    (target-llm.south east);

  \node[caption text,anchor=east] at (-2.75,0.35) {Source LLM};
  \node[caption text,anchor=west] at (2.75,0.35) {Target LLM};

  \node[
    font=\sffamily\bfseries\small,
    text=ink
  ] at (0,-1.85) {Differential\\execution};

  \MatcherIcon{matcher}{(0,-4.30)}
  \node[caption text,above=2mm of matcher] {Evidence-aware matcher};
  \draw[flow]
    (source-execution.south)
    -- (-2.0,-4.30)
    -- (matcher.west);
  \draw[flow]
    (target-execution.south)
    -- (2.0,-4.30)
    -- (matcher.east);

  \node[result] (decision) at (0,-5.55)
    {Equivalent \quad / \quad Inequivalent \quad / \quad Unclear};
  \draw[flow]
    (matcher.south) -- (decision.north);

\end{tikzpicture}%
}
\vspace{-5mm}
\caption{\textbf{Graph-grounded differential assessment:} \method uses graph evidence to guide branch-targeting inputs; blinded surrogates independently predict source and target traces,
and deterministic reconciliation returns \vequiv, \vinequiv, or \vabstain judgment.}
\label{fig:function-equivalency-workflow}
\vspace{-7mm}
\endgroup
\end{figure*}

To address these limitations, this paper studies a more constrained use of model reasoning. We propose \method, illustrated in Figure~\ref{fig:function-equivalency-workflow}, which treats functional equivalence as an \emph{evidence acquisition and selective decision} problem. \method represents the programs as a pair of typed, attributed program graphs rather than an undifferentiated context window. A structural agent can inspect only bounded slices and must cite graph nodes, relation paths, and source spans. The behavioral assessment does not compare the programs directly. Instead, one component proposes inputs from uncovered branch obligations, while separate and mutually blinded agent surrogates predict the observable trace of each program. Finally, fixed reconciliation rules distinguish evidence for equivalence, evidence for divergence, and reasons to abstain.

This design makes the distinction between \emph{assessment} and \emph{proof} important. \method cannot prove universal equivalence from finitely many queries and model predictions. Instead, it estimates a scoped relation relative to an explicit input domain and observable schema, exposes its evidence, and withholds a verdict when coverage or agreement is insufficient. This makes selective risk, especially the rate of false \vequiv decisions, a more meaningful target than raw forced-choice accuracy.

Bug patching on the other hand provides a high-value application of this assessment. Benchmarks such as SWE-bench Verified \citep{jimenez2024swebench} score candidate patches with repository tests. \method adds an orthogonal comparison against the reference behavior. A test-passing candidate that loses reference behavior is a likely evaluation false positive. A test-failing candidate that remains behaviorally equivalent is a candidate false negative caused by an overly strict or incidental test. When a patch adds behavior, we use the term \extension candidate; calling it an ``improvement'' requires task intent or human adjudication, not equivalence analysis alone.

Building on this assessment approach and its application to bug patching, the main contributions of this work are:
\begin{itemize}
  \item a formulation of scoped observational equivalence that makes the input
  domain, observable behavior, and environment assumptions explicit, and
  supports \vequiv, \vinequiv, and \vabstain judgment rather than forcing a binary
  decision;
  \item \method, a graph-grounded architecture that aligns behavioral anchors,
  derives checkable obligations, generates branch-targeting inputs, and
  independently predicts source and target traces with mutually blinded
  surrogates;
  \item an auditable decision process in which claims cite graph nodes and
  source spans, predicted mismatches identify a concrete input and observable,
  and deterministic gates prevent an LLM from validating its own evidence; and
  \item an execution-grounded evaluation on 1,200 EquiBench pairs
  \citep{li2025equibench} and 331 test-passing SWE-bench Verified patches
  \citep{jimenez2024swebench}, that confirms  216 of 230 challenged EquiBench
  labels and 94 of 109 flagged patch divergences.
\end{itemize}

\section{Related Work}

\paragraphhead{Program equivalence and differential testing.} Equivalence is undecidable in general, so practical systems restrict the language, transformation, or observation boundary. Translation validation checks a concrete compiler output against its input \citep{pnueli1998translation}; Alive2 proves refinement between LLVM transformations \citep{lopes2021alive2}. Differential testing compares multiple implementations on generated inputs \citep{mckeeman1998differential}, while DART and KLEE combine concrete and symbolic execution to discover path-sensitive failures \citep{godefroid2005dart,cadar2008klee}. These
approaches are stronger than surrogate reasoning when their execution model applies. \method targets the complementary setting in which direct execution or complete symbolic models are unavailable. We evaluate its assessments against execution whenever execution is possible.

\paragraphhead{Code representations and graph-grounded models.} Code property graphs unify syntax, control flow, and data flow for program analysis \citep{yamaguchi2014cpg}. Learned code models also benefit from structural relations: GraphCodeBERT incorporates data flow \citep{guo2021graphcodebert}, while repository systems such as RepoGraph and CodexGraph use graph retrieval to support software-engineering agents \citep{ouyang2024repograph,liu2024codexgraph}. Our goal differs from repository question answering or patch generation: we use graph structure to define comparison anchors, derive coverage obligations, restrict model queries, and make every equivalence claim traceable to program evidence.

\paragraphhead{LLM execution and code reasoning.} CRUXEval measures input/output code reasoning
\citep{gu2024cruxeval}, CodeMind studies independent execution and consistency reasoning \citep{jin2024codemind}, and recent work directly benchmarks semantic code equivalence \citep{maveli2025codeequivalence}. EquiBench collects human-written and transformed function pairs across several languages and equivalence categories \citep{li2025equibench}. These works
show both the promise and fragility of model-based execution. \method builds on this capability by combining it with repository graphs, cross-program blinding, structured observables, and selective decision rules.

\paragraphhead{Patch evaluation and selective prediction.}
SWE-bench Verified evaluates repository-level issue resolution \citep{jimenez2024swebench}, and SWE-agent demonstrates the importance of the agent--computer interface \citep{yang2024sweagent}. Tests remain the primary oracle for patch evaluation, although they can under-specify intent or reject legitimate implementations. We therefore position equivalence evidence as a second audit layer, not a replacement for tests. Within this layer, our \vabstain outcome follows the principle of selective classification: a system should trade coverage for lower conditional error instead of forcing uncertain predictions \citep{geifman2017selective}. This trade-off is especially important because a false equivalence claim can conceal a regression.
\section{Problem Formulation}
\label{sec:problem-formulation}

\paragraphhead{Scoped observational equivalence.}
We compare a source program $\sourceprog$ and a target program $\targetprog$
at aligned entry points.  The comparison is defined by an explicit assessment
specification
\[
  \Sigma=(\domain,\obs,\Gamma,\tau),
\]
where $\domain$ is the admissible input domain, $\obs$ specifies the
caller-observable behavior to compare, $\Gamma$ states assumptions about
external dependencies and state, and $\tau$ defines the resource policy for
timeouts and termination.  An element $x\in\domain$ denotes one logical
invocation, including explicit arguments and relevant initial state.  If the
programs expose different concrete APIs or data representations, the task
manifest supplies fixed adapters from this logical invocation to each side.

Under this specification, the ideal equivalence relation is
\begin{equation}
\sourceprog \equiv_{\Sigma} \targetprog, \ \forall x\in\domain:\;
\obs\!\left(\sem{\sourceprog}_{\Gamma,\tau}(x)\right)
=
\obs\!\left(\sem{\targetprog}_{\Gamma,\tau}(x)\right).
\label{eq:ideal-equivalence}
\end{equation}
The observable projection includes the return value, exception behavior,
external effects, mutations of caller-reachable objects, and termination
status.  Each enabled field has a declared comparator, such as exact equality,
tolerance-bounded numeric equality, canonicalized collection equality, or an
application-specific projection.

Equation~\eqref{eq:ideal-equivalence} makes the scope of a claim explicit.
Behavior outside $\domain$ is not automatically a counterexample; internal
implementation details matter only when exposed by $\obs$; and calls into
libraries, services, files, databases, or clocks are interpreted under
$\Gamma$.  The specification must be fixed independently of the assessment,
so it cannot be changed after observing a disagreement.

\paragraphhead{Selective assessment.}
Universal program equivalence is undecidable in general, and finite model
predictions cannot prove Equation~\eqref{eq:ideal-equivalence}.  We therefore
seek a bounded, evidence-backed decision:
\[
  h(\sourceprog,\targetprog,\Sigma)
  \in \{\vequiv,\vinequiv,\vabstain\}.
\]
\method returns this verdict together with cited evidence and an assessment
confidence score.  A \vinequiv verdict requires a supported structural contradiction or
a stable behavioral mismatch for an input in $\domain$.  A \vequiv verdict
requires coverage of all mandatory comparison obligations without a known
contradiction.  When alignment, dependency behavior, path coverage, or
surrogate predictions remain unresolved, the correct output is \vabstain
rather than a forced judgment.  Thus, \vequiv denotes support under the
declared scope and finite assessment budget, not a universal proof.

\paragraphhead{Patch-specific interpretation.}
When $\sourceprog$ is the program with a reference patch and $\targetprog$ is
the program with a candidate patch, equivalence means that the candidate
preserves the reference observables under $\Sigma$.  A supported loss or
change of required reference behavior is a \regression candidate.  Additional
candidate behavior not represented by the reference is an \extension
candidate.  Whether an extension is desirable cannot be decided from
equivalence alone; it additionally requires the issue specification, tests,
or human review.

\section{\method}
\label{sec:method}

\method has five phases: graph construction, alignment and obligation extraction, bounded structural evidence channel,  two-channel evidence acquisition, and deterministic reconciliation.  Appendix~\ref{app:schemas} gives the complete graph and observable schemas.

\subsection{Typed, attributed program-pair graph}

For each program, we construct graph  $G=(V,E,A)$, where nodes, $V$, represent repository, file, namespace, type, function, parameter, statement, variable, literal, and external-symbol entities. Directed edges, $E$ encode \textsc{contains}, \textsc{declares}, \textsc{calls}, \textsc{imports}, \textsc{inherits}, \textsc{overrides}, control-flow
successors, def--use links, reads, writes, returns, throws, and externally
visible effects.  Attributes, $A$,  retain language, type, signature, source span,
branch predicate, symbol resolution status, and parser confidence.  The graph
is deliberately loss-aware: an unresolved call is an explicit node rather
than a silently missing edge.

The graph serves two roles,  1) It compresses repository context into
behaviorally meaningful neighborhoods, and  2) it defines what the assessor has
and has not inspected.  Every query records its graph version, seed nodes,
edge types, depth or budget, returned nodes, and source spans.  Evidence can
therefore be replayed after a graph or program update.
\begin{figure*}[t]
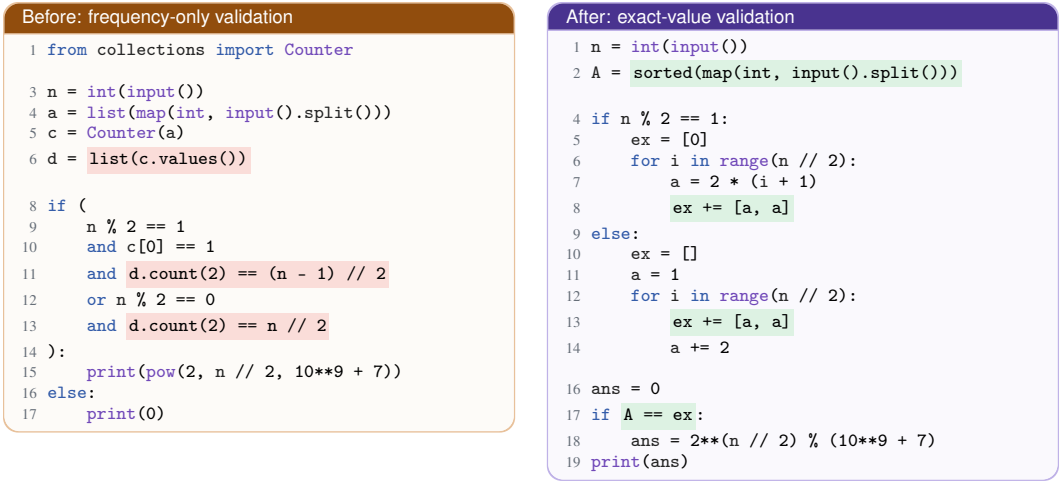

\centering
\begingroup
\setlength{\fboxsep}{1pt}

\definecolor{beforepanel}{HTML}{C96B12}
\definecolor{afterpanel}{HTML}{6647C7}
\definecolor{casegreen}{HTML}{287A42}
\definecolor{casekeyword}{HTML}{315AA8}
\definecolor{casebuiltin}{HTML}{7048B8}
\definecolor{caseink}{HTML}{242424}
\definecolor{casemuted}{HTML}{666E78}
\definecolor{casecorrecthighlight}{HTML}{DDF2E3}
\definecolor{casebughighlight}{HTML}{FADDD9}

\lstdefinestyle{combineddivergencepython}{
  language=Python,
  basicstyle=\ttfamily\fontsize{6.6pt}{7.8pt}\selectfont,
  identifierstyle=\color{caseink},
  keywordstyle=\bfseries\color{casekeyword},
  commentstyle=\itshape\color{casemuted},
  stringstyle=\color{casegreen},
  emph={Counter,int,input,list,map,pow,print,range,sorted},
  emphstyle=\color{casebuiltin},
  numbers=left,
  numberstyle=\color{casemuted}\fontsize{5.6pt}{6.5pt}\selectfont,
  numbersep=4pt,
  numberblanklines=false,
  columns=fullflexible,
  keepspaces=true,
  showstringspaces=false,
  breaklines=false,
  frame=none,
  xleftmargin=1.8em,
  escapeinside={(*@}{@*)},
  aboveskip=0pt,
  belowskip=0pt
}

\tcbset{
  codepairpanel/.style={
    enhanced jigsaw,
    listing only,
    listing engine=listings,
    colback=#1!3,
    colframe=#1!42,
    colbacktitle=#1!72!black,
    coltitle=white,
    fonttitle=\sffamily\scriptsize\mdseries,
    boxrule=0.55pt,
    arc=2mm,
    outer arc=2mm,
    boxsep=0.8pt,
    left=2pt,
    right=2pt,
    top=1.5pt,
    bottom=1.5pt,
    toptitle=1.5pt,
    bottomtitle=1.5pt,
    before skip=0pt,
    after skip=0pt
  }
}

\begin{minipage}[t]{0.485\textwidth}
  \vspace{0pt}
  \begin{tcblisting}{
    codepairpanel=beforepanel,
    title={\  \ Before: frequency-only validation},
    listing options={
      style=combineddivergencepython,
      firstnumber=1
    }
  }
from collections import Counter

n = int(input())
a = list(map(int, input().split()))
c = Counter(a)
d = (*@\colorbox{casebughighlight}{\strut\ttfamily list(c.values())}@*)

if (
    n % 2 == 1
    and c[0] == 1
    and (*@\colorbox{casebughighlight}{\strut\ttfamily d.count(2) == (n - 1) // 2}@*)
    or n % 2 == 0
    and (*@\colorbox{casebughighlight}{\strut\ttfamily d.count(2) == n // 2}@*)
):
    print(pow(2, n // 2, 10**9 + 7))
else:
    print(0)
  \end{tcblisting}
\end{minipage}
\hfill
\begin{minipage}[t]{0.485\textwidth}
  \vspace{0pt}
  \begin{tcblisting}{
    codepairpanel=afterpanel,
    title={ \  \ After: exact-value validation},
    listing options={
      style=combineddivergencepython,
      firstnumber=1
    }
  }
n = int(input())
A = (*@\colorbox{casecorrecthighlight}{\strut\ttfamily sorted(map(int, input().split()))}@*)

if n % 2 == 1:
    ex = [0]
    for i in range(n // 2):
        a = 2 * (i + 1)
        (*@\colorbox{casecorrecthighlight}{\strut\ttfamily ex += [a, a]}@*)
else:
    ex = []
    a = 1
    for i in range(n // 2):
        (*@\colorbox{casecorrecthighlight}{\strut\ttfamily ex += [a, a]}@*)
        a += 2

ans = 0
if (*@\colorbox{casecorrecthighlight}{\strut\ttfamily A == ex}@*):
    ans = 2**(n // 2) % (10**9 + 7)
print(ans)
  \end{tcblisting}
\end{minipage}

\caption{\textbf{Combined algorithm--data divergence (OJ-VA case 370):} For
\(n=4\) and \([0,0,2,2]\), multiplicity-only validation discards the
\texttt{Counter} keys and returns \(4\), whereas exact sequence validation
rejects the values and returns \(0\).}
\label{fig:verified-combined-divergence-case}
\vspace{-5mm}
\endgroup
\end{figure*}

\subsection{Anchor alignment and proof obligations}

We first align declared entry points using qualified names, signatures, type
compatibility, documentation, and call-context similarity.  We then expand
from each entry point to find behavior anchors:
\begin{itemize}
  \item public inputs, outputs, exceptions, mutations, and effectful calls;
  \item branch and loop predicates that can change an observable;
  \item transitive callees and data dependencies on an observable path; and
  \item environment assumptions, including unresolved external symbols.
\end{itemize}
Cross-program candidate anchors are scored from symbol, type, role, and local graph context.  Ambiguous many-to-many alignments remain explicit.

The aligned anchors induce a set of obligations,
$\mathcal{B}=\mathcal{B}_{\mathrm{interface}}\cup
\mathcal{B}_{\mathrm{control}}\cup
\mathcal{B}_{\mathrm{data}}\cup
\mathcal{B}_{\mathrm{effects}}\cup
\mathcal{B}_{\mathrm{environment}}$.
An obligation is not a proof theorem; it is a checkable question such as ``do both programs reject a zero denominator with the same observable exception?'', ``can this write reach the returned object?'', or ``does the target preserve the source's overflow guard?''  Each obligation has status \textsc{supported}, \textsc{contradicted}, \textsc{unresolved}, or
\textsc{out-of-scope}.

Figure~\ref{fig:verified-combined-divergence-case} illustrates why algorithmic and data obligations must be considered together: a frequency-based rewrite changes both the validation procedure and the exact values retained for the final comparison.

\subsection{Bounded structural evidence channel}

The structural analyst receives $\Sigma$, the aligned anchors, and a graph query API.  It does not receive a flattened repository.  Available operations include bounded neighborhood expansion, typed path search, callers/callees, control dependence, backward data slices, symbol lookup, and source retrieval for returned nodes.  A response is accepted into the ledger only if it contains:
\[
  e=(\text{claim},\text{polarity},\text{node IDs},
     \text{relation path},\text{source spans},\text{assumptions}).
\]
Unsupported prose is discarded.  Query depth, result count, and total tokens are bounded; hitting a bound marks the affected obligation unresolved.

The analyst compares \emph{behavioral roles}, not graph isomorphism.  Different loop forms or helper decompositions can support equivalence when their observable slices agree.  Conversely, a one-token predicate change can contradict an obligation even when the surrounding graphs align.

\subsection{Differential surrogate execution}

\paragraphhead{Branch-aware input proposals.}
The input generator converts uncovered control and data obligations into typed input  constraints.  It proposes ordinary cases, boundary values, values immediately around branch thresholds, aliasing and mutation cases, empty and singleton structures, exceptional inputs permitted by $\domain$, and combinations that distinguish mismatched predicates.  Each proposal identifies the targeted obligation and includes a rationale. Proposals are de-duplicated using canonical typed representations rather than string equality.

\paragraphhead{Blinded simulation.}
For every $x\in\tests$, separate source and target surrogates receive only one program slice, the same input, the observable schema, and relevant environment contracts.  They predict a structured trace
\[
  T_P(x)=(r,e,s,m,t,\pi),
\]
where $r$ is the return value, $e$ the exception, $s$ external side effects, $m$ reachable-parameter mutations, $t$ termination status, and $\pi$ a stepwise trace anchored to graph nodes.  Neither surrogate sees the other program or prediction.  This prevents the common failure mode in which a joint judge assumes agreement and bends both explanations toward that prior.

Each prediction undergoes a self-verification pass that checks type consistency, branch feasibility, trace-to-graph citations, and agreement between the claimed path and final observables.  We sample repeated predictions only where the initial trace is low-confidence or
self-inconsistent.  Persistent within-program disagreement marks the case unresolved rather than selecting the majority answer without qualification.

\paragraphhead{Observable comparison.}
A non-generative comparator applies the field-specific rules in $\obs$.  It returns \textsc{match}, \textsc{mismatch}, or \textsc{unknown} per field. A mismatch record contains the concrete input, the first divergent observable, both predicted values, cited trace nodes, and any assumptions.  The comparator does not infer intent and cannot relabel an extension as an improvement.

\subsection{Deterministic reconciliation and confidence}
\label{sec:reconciliation}

Reconciliation is ordered to privilege identifiable failures over broad
similarity:
\begin{enumerate}[leftmargin=*,itemsep=2pt,topsep=3pt]
  \item \textbf{Validity gate.}  If required graphs, entry-point alignments, or   observable comparators are missing, return \vabstain.
  \item \textbf{Contradiction gate.}  Return \vinequiv when a structurally   supported interface contradiction or a stable behavioral mismatch lies   inside $\domain$ and does not depend on an unresolved environment fact.
  \item \textbf{Conflict gate.}  If the structural and behavioral channels   disagree, or repeated simulations are unstable, return \vabstain and retain   the conflict.
  \item \textbf{Coverage gate.}  Return \vequiv only if every mandatory   obligation is supported or behaviorally exercised, no contradiction exists,   and aggregate confidence exceeds the acceptance threshold.
  \item Otherwise return \vabstain with the remaining obligations.
\end{enumerate}

Confidence is computed from observable-comparison stability, obligation coverage, graph resolution quality, and evidence completeness.  It is calibrated on a development partition and frozen before test evaluation. The threshold is swept for risk--coverage analysis; one operating point is selected from a preregistered upper bound on false-equivalence risk.  No LLM is
asked to override these gates.

\section{Benchmarking Results}
\label{sec:experiments}

We evaluate the reliability and utility of \method through four questions:
\begin{itemize}[leftmargin=*,itemsep=1pt,topsep=2pt]
  \item \textbf{RQ1:} Are disagreements with benchmark labels genuine label
  errors or \method errors?
  \item \textbf{RQ2:} Can \method find execution-confirmed behavioral
  divergences in test-passing patches?
  \item \textbf{RQ3:} Do generated witnesses turn predicted discrepancies into
  execution-confirmed divergences?
  \item \textbf{RQ4:} How does graph context help localize observable
  differences and the paths that reach them?
\end{itemize}
We study RQ1 and RQ3 on EquiBench
(Section~\ref{sec:results:label-quality}) and RQ2 on SWE-bench Verified
(Section~\ref{sec:results:swebench}).  We address RQ4 through an
execution-verified SWE-bench Verified case study (Section~\ref{sec:results:surrogate});
a quantitative graph ablation remains future work.

\paragraphhead{Model configuration.}
All model-backed components of \method in the reported experiments used
Claude Opus~4.6 with a 32{,}000-token thinking budget and prompt caching
enabled.
\subsection{Benchmarks}
\label{sec:results:benchmarks}

We use EquiBench \citep{li2025equibench} to audit labeled program pairs and
SWE-bench Verified \citep{jimenez2024swebench} to audit test-passing patches.
Our EquiBench evaluation covers the Python OJ\_A, OJ\_V, and OJ\_VA subsets;
Appendix Table~\ref{tab:results:benchmarks} reports their composition.  We
exclude the C/DCE, CUDA, and x86-64 subsets because our current graph front end
does not support them.  Section~\ref{sec:results:swebench} describes the
SWE-bench Verified construction and execution audit.

\subsection{Evaluation metrics}
\label{sec:results:metrics}

Published labels can be wrong, so accuracy against them conflates method and
label errors.  We instead report three execution-grounded metrics:
\begin{itemize}[leftmargin=*,itemsep=1pt,topsep=2pt]
  \item \textbf{Disagreement Rate (DR):} fraction of pairs where \method disagrees with the label;
  \vabstain counts because it does not confirm the label.
  \item \textbf{Verified Genuine (VG):} fraction of disagreements confirmed as label errors or execution-confirmed divergence.
  \item \textbf{Execution-confirmed Divergence (ECD):} fraction of all
  evaluated cases for which \method identifies a divergence that direct
  execution confirms.
\end{itemize}

Let $P$ be all pairs, $S$ the disagreements or say the inequivalent flags, and
$D(S)=\{i\in S:\mathrm{div}(i)\}$ the execution-confirmed divergences in set
$S$.  Then
\begin{equation}
\mathrm{DR}=\frac{|S|}{|P|},\quad
\mathrm{VG}=\frac{|D(S)|}{|S|},\quad
\mathrm{ECD}=\frac{|D(S)|}{|P|}, \quad
\label{eq: metrics}
\end{equation}

\subsection{Label quality verification}
\label{sec:results:label-quality}

To answer RQ1, and as Section~\ref{sec:results:metrics} anticipated,
ground-truth labels can be unreliable.  For every pair where \method disagrees
with the provided label, we apply the asymmetric verification behind the
verified-as-genuine metric:
\begin{itemize}[leftmargin=*,itemsep=1pt,topsep=2pt]
  \item \textbf{Label equivalent, prediction inequivalent.}  We produce at
  least one concrete input on which the ``before'' and ``after'' programs yield
  different outputs and execute it in a real environment to confirm the
  divergence.
  \item \textbf{Label inequivalent, prediction equivalent.}  We attempt to
  reproduce the labeled divergence by executing both programs on inputs derived
  from \method's reasoning trace.  If no input reproduces it, the evidence is
  inconclusive; we retain the original label and count the pair as a method
  failure.
\end{itemize}

Table~\ref{tab:results:label-quality} combines the EquiBench label-quality
results with the SWE-bench Verified execution audit. OJ\_A and OJ\_VA show high disagreement rates (29\% and 27.25\%) paired with very high verification rates (95.7\% and 96.3\%), indicating that the vast majority of disagreements stem from incorrect ground-truth labels rather than
from \method errors. 
\begin{figure*}[t]
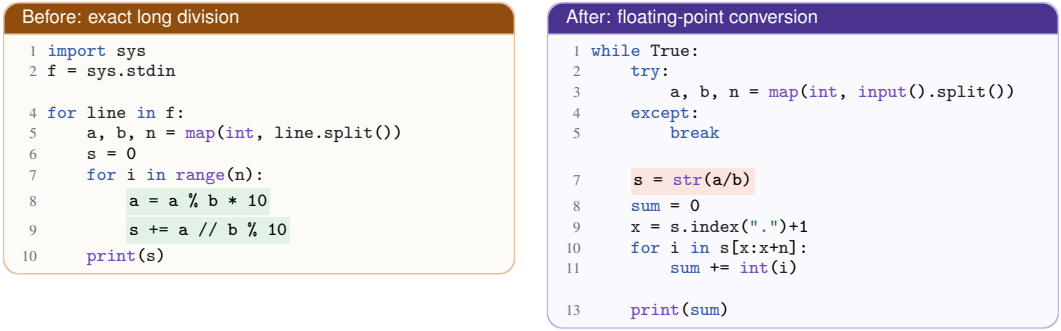

\centering
\begingroup
\setlength{\fboxsep}{1pt}

\definecolor{beforepanel}{HTML}{C96B12}
\definecolor{afterpanel}{HTML}{6647C7}
\definecolor{casegreen}{HTML}{287A42}
\definecolor{casekeyword}{HTML}{315AA8}
\definecolor{casebuiltin}{HTML}{7048B8}
\definecolor{caseink}{HTML}{242424}
\definecolor{casemuted}{HTML}{666E78}
\definecolor{caseexacthighlight}{HTML}{E2F3E6}
\definecolor{casefloathighlight}{HTML}{FCE4E1}

\lstdefinestyle{divergencepython}{
  language=Python,
  basicstyle=\ttfamily\fontsize{6.6pt}{7.8pt}\selectfont,
  identifierstyle=\color{caseink},
  keywordstyle=\bfseries\color{casekeyword},
  commentstyle=\itshape\color{casemuted},
  stringstyle=\color{casegreen},
  emph={int,input,map,print,range,str},
  emphstyle=\color{casebuiltin},
  numbers=left,
  numberstyle=\color{casemuted}\fontsize{5.6pt}{6.5pt}\selectfont,
  numbersep=4pt,
  numberblanklines=false,
  columns=fullflexible,
  keepspaces=true,
  showstringspaces=false,
  breaklines=false,
  frame=none,
  xleftmargin=1.8em,
  escapeinside={(*@}{@*)},
  aboveskip=0pt,
  belowskip=0pt
}

\tcbset{
  codepairpanel/.style={
    enhanced jigsaw,
    listing only,
    listing engine=listings,
    colback=#1!3,
    colframe=#1!42,
    colbacktitle=#1!72!black,
    coltitle=white,
    fonttitle=\sffamily\scriptsize\mdseries,
    boxrule=0.55pt,
    arc=2mm,
    outer arc=2mm,
    boxsep=0.8pt,
    left=2pt,
    right=2pt,
    top=1.5pt,
    bottom=1.5pt,
    toptitle=1.5pt,
    bottomtitle=1.5pt,
    before skip=0pt,
    after skip=0pt
  }
}

\begin{minipage}[t]{0.485\textwidth}
  \vspace{0pt}
  \begin{tcblisting}{
    codepairpanel=beforepanel,
    title={\  \ Before: exact long division},
    listing options={
      style=divergencepython,
      firstnumber=1,
      lastline=10
    }
  }
import sys
f = sys.stdin

for line in f:
    a, b, n = map(int, line.split())
    s = 0
    for i in range(n):
        (*@\colorbox{caseexacthighlight}{\strut\ttfamily a = a \% b * 10}@*)
        (*@\colorbox{caseexacthighlight}{\strut\ttfamily s += a // b \% 10}@*)
    print(s)
  \end{tcblisting}
\end{minipage}
\hfill
\begin{minipage}[t]{0.485\textwidth}
  \vspace{0pt}
  \begin{tcblisting}{
    codepairpanel=afterpanel,
    title={\  \ After: floating-point conversion},
    listing options={
      style=divergencepython,
      firstnumber=1,
      lastline=13
    }
  }
while True:
    try:
        a, b, n = map(int, input().split())
    except:
        break

    (*@\colorbox{casefloathighlight}{\strut\ttfamily s = \textcolor{casebuiltin}{str}(a/b)}@*)
    sum = 0
    x = s.index(".")+1
    for i in s[x:x+n]:
        sum += int(i)

    print(sum)
  \end{tcblisting}
\end{minipage}

\caption{\textbf{Algorithmic divergence from numeric semantics (OJ-A case
10):} For a benchmark-labeled equivalent pair, \(a=10\), \(b=3\), and \(n=16\)
produce \(48\) versus \(50\): replacing exact long division with rounded
floating-point conversion changes the digit sum.}
\label{fig:verified-divergence-case}

\endgroup
\end{figure*}

\begin{table}[!ht]
\caption{\textbf{Execution-grounded audit on EquiBench and SWE-bench Verified.} ECD
$=$ execution-confirmed divergences / evaluated cases as defined in \ref{eq: metrics}.  On SWE-bench Verified we
evaluate the 331 instances resolved (i.e. every patch passes its tests) by mini-SWE-agent with GPT-OSS-120B.}
\label{tab:results:label-quality}
\label{tab:results:swebench}
\centering
\small
\setlength{\tabcolsep}{2.5pt}
\begin{tabular}{lccccc}
\toprule
Audit Set & Evaluated & \shortstack{Disagreement Rate}
  & \shortstack{Verified Genuine}
  & \shortstack{ECD}\\
\midrule
OJ\_A  & 400 & 116 (29\%)    & 111 (95.7\%) & 27.8\%\\
OJ\_VA & 400 & 109 (27.25\%) & 105 (96.3\%) & 26.3\%\\
OJ\_V  & 400 & 5 (1.25\%)    & 0 (0\%)      & 0\%\\
\cmidrule(lr){1-5}
EquiBench Total & 1,200 & 230 (19.2\%) & 216 (93.9\%) & \textbf{\EquiBenchECD}\\
\midrule
SWE-bench Verified
  & 331 & 109 (32.9\%) & 94 (86.2\%) & \textbf{\SWEBenchECD}\\
\bottomrule
\end{tabular}
\vspace{-3mm}
\end{table}

\begin{figure*}[t]
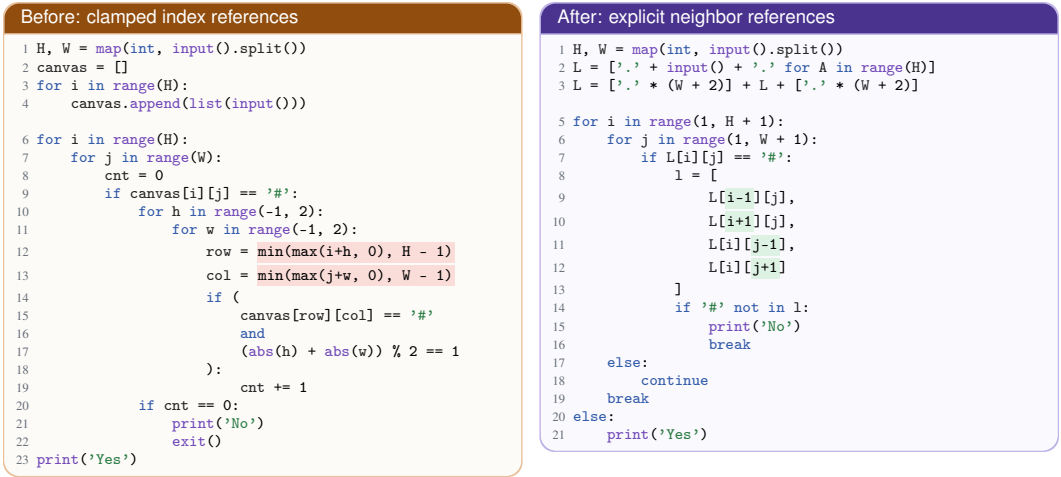

\centering
\begingroup
\setlength{\fboxsep}{0.5pt}

\definecolor{beforepanel}{HTML}{C96B12}
\definecolor{afterpanel}{HTML}{6647C7}
\definecolor{casegreen}{HTML}{287A42}
\definecolor{casekeyword}{HTML}{315AA8}
\definecolor{casebuiltin}{HTML}{7048B8}
\definecolor{caseink}{HTML}{242424}
\definecolor{casemuted}{HTML}{666E78}
\definecolor{casecorrecthighlight}{HTML}{DDF2E3}
\definecolor{casebughighlight}{HTML}{FADDD9}

\lstdefinestyle{variabledivergencepython}{
  language=Python,
  basicstyle=\ttfamily\fontsize{5.7pt}{6.8pt}\selectfont,
  identifierstyle=\color{caseink},
  keywordstyle=\bfseries\color{casekeyword},
  commentstyle=\itshape\color{casemuted},
  stringstyle=\color{casegreen},
  emph={abs,append,exit,input,len,list,map,max,min,print,range},
  emphstyle=\color{casebuiltin},
  numbers=left,
  numberstyle=\color{casemuted}\fontsize{4.8pt}{5.7pt}\selectfont,
  numbersep=3pt,
  numberblanklines=false,
  columns=fullflexible,
  keepspaces=true,
  showstringspaces=false,
  breaklines=false,
  frame=none,
  xleftmargin=1.55em,
  escapeinside={(*@}{@*)},
  aboveskip=0pt,
  belowskip=0pt
}

\tcbset{
  codepairpanel/.style={
    enhanced jigsaw,
    listing only,
    listing engine=listings,
    colback=#1!3,
    colframe=#1!42,
    colbacktitle=#1!72!black,
    coltitle=white,
    fonttitle=\sffamily\scriptsize\mdseries,
    boxrule=0.55pt,
    arc=2mm,
    outer arc=2mm,
    boxsep=0.8pt,
    left=1.5pt,
    right=1.5pt,
    top=1.5pt,
    bottom=1.5pt,
    toptitle=1.5pt,
    bottomtitle=1.5pt,
    before skip=0pt,
    after skip=0pt
  }
}

\begin{minipage}[t]{0.492\textwidth}
  \vspace{0pt}
  \begin{tcblisting}{
    codepairpanel=beforepanel,
    title={\  \ Before: clamped index references},
    listing options={
      style=variabledivergencepython,
      firstnumber=1
    }
  }
H, W = map(int, input().split())
canvas = []
for i in range(H):
    canvas.append(list(input()))

for i in range(H):
    for j in range(W):
        cnt = 0
        if canvas[i][j] == '#':
            for h in range(-1, 2):
                for w in range(-1, 2):
                    row = (*@\colorbox{casebughighlight}{\strut\ttfamily min(max(i+h, 0), H - 1)}@*)
                    col = (*@\colorbox{casebughighlight}{\strut\ttfamily min(max(j+w, 0), W - 1)}@*)
                    if (
                        canvas[row][col] == '#'
                        and
                        (abs(h) + abs(w)) % 2 == 1
                    ):
                        cnt += 1
            if cnt == 0:
                print('No')
                exit()
print('Yes')
  \end{tcblisting}
\end{minipage}
\hfill
\begin{minipage}[t]{0.492\textwidth}
  \vspace{0pt}
  \begin{tcblisting}{
    codepairpanel=afterpanel,
    title={\  \  After: explicit neighbor references},
    listing options={
      style=variabledivergencepython,
      firstnumber=1
    }
  }
H, W = map(int, input().split())
L = ['.' + input() + '.' for A in range(H)]
L = ['.' * (W + 2)] + L + ['.' * (W + 2)]

for i in range(1, H + 1):
    for j in range(1, W + 1):
        if L[i][j] == '#':
            l = [
                L[(*@\colorbox{casecorrecthighlight}{\strut\ttfamily i-1}@*)][j],
                L[(*@\colorbox{casecorrecthighlight}{\strut\ttfamily i+1}@*)][j],
                L[i][(*@\colorbox{casecorrecthighlight}{\strut\ttfamily j-1}@*)],
                L[i][(*@\colorbox{casecorrecthighlight}{\strut\ttfamily j+1}@*)]
            ]
            if '#' not in l:
                print('No')
                break
    else:
        continue
    break
else:
    print('Yes')
  \end{tcblisting}
\end{minipage}

\caption{\textbf{Boundary-reference divergence (OJ-VA case 288):} On
\(\{\texttt{\#..},\texttt{...},\texttt{...}\}\), boundary clamping makes the
corner cell count itself as a neighbor and output \texttt{Yes}, whereas padded
neighbor references correctly output \texttt{No}.}
\label{fig:verified-variable-divergence-case}
\vspace{-5mm}
\endgroup
\end{figure*}

 Again, in each confirmed case we found at least one
concrete input on which the ``before'' and ``after'' programs produce different
outputs and executed the code to validate the agent's reasoning.  OJ\_V, by
contrast, has a low disagreement rate (1.25\%) and none of its disagreements
were genuine mislabels.  Across all 1{,}200 evaluated EquiBench
pairs, ECD is \EquiBenchECD($216/1{,}200$): at least this fraction of pairs
carries a published label contradicted by an executed counterexample.

Figures~\ref{fig:verified-combined-divergence-case}, \ref{fig:verified-divergence-case} and  \ref{fig:verified-variable-divergence-case},   give representative
execution-verified disagreements, including combined algorithm with variable divergences algorithmic divergences, and variable-level divergences.

\subsection{SWE-bench Verified regression audit}
\label{sec:results:swebench}

We generate candidate fixes for the 500 SWE-bench Verified issues
\citep{jimenez2024swebench} with mini-SWE-agent
\citep{yang2024sweagent}, retaining 331 patches that pass all tests.  We compare
each with its reference patch, and \method flags 109 (32.9\%) as inequivalent.
For each flag, its reasoning trace guides an adversarial input that we execute
against both revisions in separate subprocesses.

Execution confirms 94 flags (86.2\%); 14 do not reproduce because the change is
compensated elsewhere, unreachable, or observably equivalent, and one remains
inconclusive.  Since all 331 candidates passed the tests, the 94 confirmed
divergences give an execution-confirmed divergence yield of $94/331=28.4\%$
(Table~\ref{tab:results:swebench}).

\subsection{Value of surrogate execution and code graph}
\label{sec:results:surrogate}

EquiBench's self-contained programs make each generated witness directly checkable.  For a challenged pair, the branch-aware generator targets an unresolved control or data obligation, and blinded surrogates predict each program's trace independently.  Executing the proposed input confirmed every disagreement counted as genuine, converting a predicted discrepancy into a
reproducible counterexample.  Figures~\ref{fig:verified-combined-divergence-case} , \ref{fig:verified-divergence-case}, and  \ref{fig:verified-variable-divergence-case} illustrate  combined- boundary and algorithm,  boundary-reference and algorithmic failures.  This establishes practical value; causal attribution requires a matched ablation.

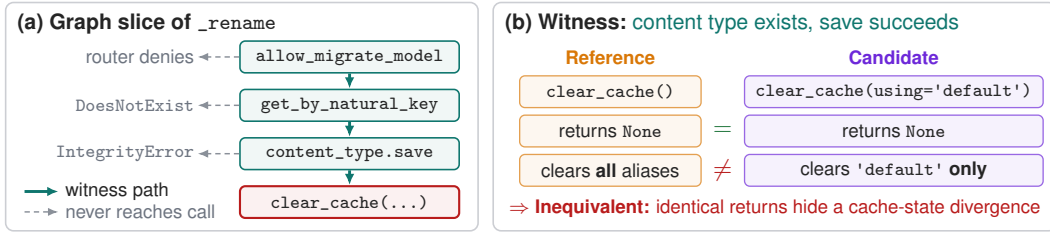
\begin{figure}[!h]
\centering
\begingroup

\definecolor{caseink}{HTML}{252525}
\definecolor{casemuted}{HTML}{6D7480}
\definecolor{casematch}{HTML}{287A42}

\begin{tikzpicture}[
  x=1cm,
  y=1cm,
  font=\sffamily\scriptsize,
  every node/.style={text=caseink},
  panel/.style={
    draw=casemuted!35,
    fill=white,
    rounded corners=3pt,
    line width=0.6pt
  },
  panel title/.style={
    anchor=north west,
    font=\sffamily\footnotesize\bfseries,
    inner sep=0pt
  },
  slice/.style={
    draw=graphteal,
    fill=graphteal!6,
    rounded corners=2pt,
    line width=0.6pt,
    minimum width=29mm,
    minimum height=4.2mm,
    inner sep=1.5pt,
    font=\ttfamily\scriptsize
  },
  effect/.style={
    slice,
    draw=warningred,
    fill=warningred!6,
    line width=0.8pt
  },
  exit/.style={
    anchor=east,
    text=casemuted,
    inner sep=1pt
  },
  witness/.style={
    -{Latex[length=1.5mm,width=1.3mm]},
    draw=graphteal,
    line width=1.1pt
  },
  pruned/.style={
    -{Latex[length=1.4mm,width=1.1mm]},
    draw=casemuted!80,
    line width=0.55pt,
    dash pattern=on 2pt off 1.5pt
  },
  cell/.style={
    rounded corners=2pt,
    line width=0.5pt,
    minimum height=4.2mm,
    inner xsep=3pt,
    inner ysep=1pt
  },
  ref cell/.style={
    cell,
    draw=sourceorange!70,
    fill=sourceorange!6,
    minimum width=24mm
  },
  cand cell/.style={
    cell,
    draw=targetviolet!70,
    fill=targetviolet!6,
    minimum width=39.5mm
  },
  relation/.style={
    font=\sffamily\small\bfseries,
    inner sep=0pt
  }
]

\draw[panel] (0,-0.22) rectangle (6.25,2.80);
\node[panel title] at (0.15,2.66) {(a) Graph slice of \texttt{\_rename}};

\node[slice]  (router) at (4.55,2.08) {allow\_migrate\_model};
\node[slice]  (lookup) at (4.55,1.44) {get\_by\_natural\_key};
\node[slice]  (save)   at (4.55,0.80) {content\_type.save};
\node[effect] (call)   at (4.55,0.16) {clear\_cache(\ldots)};

\draw[witness] (router) -- (lookup);
\draw[witness] (lookup) -- (save);
\draw[witness] (save)   -- (call);

\foreach \guard/\why in {
  router/router denies,
  lookup/\texttt{DoesNotExist},
  save/\texttt{IntegrityError}%
} {
  \draw[pruned] (\guard.west) -- ++(-0.55,0) node[exit] {\why};
}

\draw[witness] (0.25,0.30) -- ++(0.45,0);
\node[anchor=west,inner sep=1pt] at (0.75,0.30) {witness path};
\draw[pruned] (0.25,0.03) -- ++(0.45,0);
\node[anchor=west,inner sep=1pt,text=casemuted] at (0.75,0.03)
  {never reaches call};

\draw[panel] (6.45,-0.22) rectangle (13.9,2.80);
\node[panel title] at (6.60,2.66)
  {(b) Witness:\ \textcolor{graphteal}{\mdseries content type exists, save
  succeeds}};

\node[font=\sffamily\scriptsize\bfseries,text=sourceorange]
  at (8.00,2.08) {Reference};
\node[font=\sffamily\scriptsize\bfseries,text=targetviolet]
  at (11.75,2.08) {Candidate};

\node[ref cell]  (ref-call)  at (8.00,1.62) {\texttt{clear\_cache()}};
\node[cand cell] (cand-call) at (11.75,1.62)
  {\texttt{clear\_cache(using=\textquotesingle default\textquotesingle)}};

\node[ref cell]  (ref-ret)  at (8.00,1.10) {returns \texttt{None}};
\node[cand cell] (cand-ret) at (11.75,1.10) {returns \texttt{None}};
\node[relation,text=casematch]
  at ($(ref-ret.east)!0.5!(cand-ret.west)$) {\(=\)};

\node[ref cell]  (ref-eff)  at (8.00,0.58) {clears \textbf{all} aliases};
\node[cand cell] (cand-eff) at (11.75,0.58)
  {clears \texttt{\textquotesingle default\textquotesingle} \textbf{only}};
\node[relation,text=warningred]
  at ($(ref-eff.east)!0.5!(cand-eff.west)$) {\(\neq\)};

\node[anchor=center,inner sep=0pt,text=warningred] at (10.175,0.08)
  {\(\Rightarrow\) \textbf{Inequivalent:} identical returns hide a cache-state
  divergence};

\end{tikzpicture}

\caption{\textbf{Graph-guided divergence in \texttt{django-11066}.} (a) Only
guard-satisfying inputs reach \texttt{clear\_cache}. (b) Both revisions return
\texttt{None}, but the candidate clears only the \texttt{default} alias.}
\label{fig:django-graph-case}

\endgroup
\vspace{-5mm}
\end{figure}

Figure~\ref{fig:django-graph-case} shows why graph context matters when return values match.  The slice links \texttt{RenameContentType.\_rename} to the effectful \texttt{clear\_cache} call and its router, lookup, and save guards.
This guides generation of an existing content type with a successful save,
reaching \texttt{clear\_cache()} in the reference and
\texttt{clear\_cache(using='default')} in the candidate.  Inputs taking the
not-found or integrity-error branches never reach the call.  The graph thus
provides both the observable effect and its reachability conditions, although
quantitative attribution still requires an ablation.

\section{Discussion and Conclusion}
\label{sec:discussion}

\paragraphhead{Findings.}
The evidence supports a focused conclusion: \method is an effective second
oracle for finding behavioral differences missed by benchmark labels and unit
tests.  Direct execution confirmed 216 of 230 EquiBench disagreements and 94
of 109 divergences flagged among test-passing SWE-bench Verified patches.  As a
fraction of all evaluated cases, this gives an ECD of
\EquiBenchECD($216/1{,}200$) on EquiBench and \SWEBenchECD($94/331$) on
SWE-bench Verified, where every evaluated patch had already passed its tests.  On
SWE-bench Verified, more than one quarter of test-passing patches therefore hide an
execution-confirmed divergence from the reference patch.


\paragraphhead{Scope and limitation.}
\method is a selective audit layer, not a formal verifier.  A \vinequiv verdict
prioritizes a concrete candidate counterexample, \vequiv requires covered
obligations without a stable contradiction, and \vabstain exposes missing
evidence. Surrogate traces only propose witnesses: every divergence we report
was reproduced by direct execution, and it holds only within the declared input
domain, observables, and environment contract. Incomplete graphs, unresolved
symbols, and surrogate errors can still lead \method to miss a divergence or
abstain. 

\paragraphhead{Conclusion.}
By combining graph-grounded obligations, blinded surrogate execution, and
deterministic reconciliation, \method turns model reasoning into concrete,
inspectable evidence while keeping uncertainty explicit.  The results show
that this audit layer can expose substantial blind spots in labels and unit
tests while direct execution prevents unsupported relabeling.  The next
priorities are matched-budget component ablations and execution audits of
unflagged cases.



\bibliographystyle{plainnat}
\bibliography{references}


\clearpage
\appendix
\section{Artifact and Protocol Details}
\label{app:protocol}

\subsection{End-to-end assessment procedure}

The assessment follows the same ordered procedure for every pair.  Run-level
settings are recorded before evaluation and held fixed within each benchmark
suite.  Capitalized returns are machine-readable verdicts with typed reason
codes.

\begin{enumerate}[leftmargin=*,itemsep=2pt]
  \item Parse $\sourceprog$ and $\targetprog$ with language-specific front ends.
  Build $\sourcegraph$ and $\targetgraph$ and record unresolved symbols.
  If either required entry point cannot be represented, return
  \textsc{Abstain:GraphFailure}.
  \item Load $\Sigma=(\domain,\obs,\Gamma,\tau)$ from the task manifest.  Check
  that every observable has a comparator.  Otherwise return
  \textsc{Abstain:IncompleteSpecification}.
  \item Align entry points and behavior anchors.  If a mandatory alignment is
  ambiguous after the query budget, return
  \textsc{Abstain:Alignment}.
  \item Extract interface, control, data, effect, and environment obligations.
  Run the structural analyst through the bounded graph API.  Reject ledger
  entries without valid node IDs, paths, and spans.
  \item Generate typed inputs targeting unresolved obligations.  Canonicalize
  and deduplicate them.  Inputs outside $\domain$ may be retained for a
  separate robustness analysis but cannot determine the primary verdict.
  \item Independently simulate each program on each input. Apply the
self-verification pass described in Appendix~C.4 to validate the trace
schema, branch feasibility, types, and citations. Re-sample only cases
selected by the predefined instability rule.
  \item Compare observable fields with deterministic comparators.  Record the
  first mismatch and all unknown fields.
  \item Apply the ordered gates in Section~\ref{sec:reconciliation}.  Emit the
  verdict, confidence, obligation matrix, evidence ledger, token/query usage,
  and all abstention reasons.
\end{enumerate}

\subsection{Run and audit records}

A reproducible run record contains the graph front end and language version,
model identifier and access date, decoding policy, graph-query and token
budgets, maximum generated inputs, simulation repetition rule, and direct
execution timeout and resource policy.  The corresponding pair record contains
immutable source and target identifiers; the scoped specification $\Sigma$;
entry-point alignments; graph-resolution warnings; extracted obligations;
generated inputs; source and target traces; comparator outputs; the first
mismatching observable, if any; cited graph evidence; and the final verdict
with every abstention or unresolved reason.  Keeping configuration, proposed
evidence, and observed outcomes separate prevents an incomplete run from being
mistaken for an equivalent result.

\section{Graph and Evidence Schemas}
\label{app:schemas}

\subsection{Evidence-ledger record}

Each record is serialized as:
\begin{verbatim}
{
  "obligation_id": "...",
  "channel": "structural | source_sim | target_sim | comparator",
  "claim": "...",
  "polarity": "support | contradict | unresolved",
  "graph_version": "...",
  "node_ids": ["..."],
  "relation_path": ["CALLS", "CONTROL_DEPENDS", "..."],
  "source_spans": [{"file": "...", "start": ..., "end": ...}],
  "input_id": "...",
  "observable_field": "...",
  "assumptions": ["..."],
  "confidence_components": {"...": ...}
}
\end{verbatim}

The artifact validator rejects dangling node IDs, spans outside the frozen
source snapshot, relation paths absent from the graph, and observable fields
not declared in $\obs$.

\begin{table}[!ht]
\caption{Minimum typed program-graph schema.  Front ends may add
language-specific nodes and relations, but cannot silently omit unresolved
entities.}
\label{tab:graph-schema}
\centering
\small
\begin{tabularx}{\linewidth}{L{0.18\linewidth} L{0.29\linewidth} X}
\toprule
Family & Required values & Core attributes\\
\midrule
Nodes & repository, file, namespace, type, function, parameter, statement,
variable, literal, external symbol & stable ID, language, qualified name,
source span, parser confidence\\
Structural edges & contains, declares, imports, inherits, overrides & source,
target, resolution status\\
Behavioral edges & calls, CFG successor, control-depends, defines, uses,
reads, writes, returns, throws, mutates, emits effect & source, target,
guard/label, static/dynamic confidence\\
Function attributes & signature, visibility, modifiers, return type, effect
summary & normalized and original forms\\
Statement attributes & syntax kind, branch predicate, constants, operators &
source text hash and span\\
External attributes & package/module, version constraint, contract source &
resolved, unresolved, or assumed\\
\bottomrule
\end{tabularx}
\end{table}

\begin{table}[!ht]
\caption{Observable fields and default comparison behavior.  Every task
manifest explicitly enables, disables, or overrides each field.}
\label{tab:observable-schema}
\centering
\small
\begin{tabularx}{\linewidth}{L{0.20\linewidth} L{0.34\linewidth} X}
\toprule
Field & Representation & Default comparator\\
\midrule
Return & typed, canonical value & exact after type-aware canonicalization;
declared tolerance for floating point\\
Exception & type, message projection, phase & exact type; configurable message
projection\\
External effects & ordered events over files, network, database, logs,
processes, callbacks & order-sensitive unless specification declares
commutativity\\
Input mutation & before/after reachable object projection & structural
equality with alias identities retained\\
Persistent state & selected state projection & task-defined equality\\
Termination & normal, exception, timeout, divergence & exact class under
resource policy $\tau$\\
\bottomrule
\end{tabularx}
\end{table}

\section{Agent Contracts and Information Boundaries}
\label{app:prompts}

The implementation may phrase prompts differently across model providers, but
the following role separation, information boundaries, and machine-readable
outputs are invariant.  In particular, a blinded surrogate never sees the
other program or its prediction, and no model may override deterministic
reconciliation.

\paragraphhead{Model-use scope.}
Models are used to analyze graph-grounded obligations, propose discriminating
inputs, predict one program's trace at a time, and check trace consistency.
They do not directly certify a benchmark label or amend it: deterministic
comparators produce the method verdict, and direct execution supplies every
reported verification result.  The run record binds each model output to the
exact model identifier, provider revision or access date, and decoding
configuration.

\subsection{Structural analyst}

\begin{quote}\small
\textbf{Role:} assess one comparison obligation using only the supplied graph
query API.  Do not claim whole-program equivalence.\\
\textbf{Inputs:} scoped specification, aligned anchors, unresolved obligation,
remaining query budget.\\
\textbf{Requirements:} query before concluding; distinguish code fact from
environment assumption; cite every supporting node, path, and source span;
mark missing or ambiguous relations unresolved.\\
\textbf{Output:} obligation ID, status
\{\textsc{supported}, \textsc{contradicted}, \textsc{unresolved},
\textsc{out-of-scope}\}, concise claim, evidence records, assumptions, and
suggested discriminating inputs.
\end{quote}

\subsection{Branch-aware input generator}

\begin{quote}\small
\textbf{Role:} propose valid inputs likely to distinguish the programs or
cover an unresolved obligation.\\
\textbf{Inputs:} input types and preconditions, aligned branch predicates,
constants and operators, data dependencies, already covered cases.\\
\textbf{Requirements:} include ordinary and boundary values; target both sides
of a predicate; preserve cross-parameter constraints; do not invent invalid
objects; identify the targeted obligation; do not predict either output.\\
\textbf{Output:} canonical typed input, construction steps, targeted
obligation, relevant graph nodes, expected path constraints, and validity
rationale.
\end{quote}

\subsection{Blinded execution surrogate}

\begin{quote}\small
\textbf{Role:} predict the observable execution of exactly one program on the
given input.  You are not told whether the program is the source or target and
you do not see another program's prediction.\\
\textbf{Inputs:} one program slice, graph neighborhood, input, environment
contracts, resource policy, and observable schema.\\
\textbf{Requirements:} trace values and control decisions step by step; cite
nodes for each decisive step; report unresolved external behavior rather than
guessing; include mutations and side effects even when the return value
matches; check the final value against the trace.\\
\textbf{Output:} trace events, return, exception, effects, mutations,
termination, unresolved assumptions, citations, and fieldwise confidence.
\end{quote}

\subsection{Trace verifier}

\begin{quote}\small
\textbf{Role:} find internal inconsistencies in a proposed trace, not rewrite
it into a preferred answer.\\
\textbf{Checks:} input and intermediate types; branch feasibility; definition
before use; called-function contract; trace-to-node citation; final
observable; missed mutation/effect; dependence on an unresolved symbol.\\
\textbf{Output:} \textsc{valid}, \textsc{invalid}, or \textsc{uncertain}, with
localized errors.  An invalid trace is re-simulated without revealing a
prediction from the other program.
\end{quote}

\section{Benchmark and Execution-Verification Details}
\label{app:verification}

\begin{table}[ht]
\caption{EquiBench subsets used in the study.  Equiv./Inequiv.\ give the
published class balance.}
\label{tab:results:benchmarks}
\centering
\small
\begin{tabular}{llcccl}
\toprule
Dataset & Language & Pairs & Equiv. & Inequiv. & Labeling source\\
\midrule
OJ\_A  & Python & 400 & 200 & 200 & EquiBench \citep{li2025equibench}\\
OJ\_V  & Python & 400 & 200 & 200 & EquiBench \citep{li2025equibench}\\
OJ\_VA & Python & 400 & 200 & 200 & EquiBench \citep{li2025equibench}\\
\bottomrule
\end{tabular}
\end{table}

\subsection{Pair manifest}

Every comparison is represented by a manifest containing stable source and
target revisions; entry points; $\domain$, $\obs$, $\Gamma$, and $\tau$;
expected language/toolchain; provenance; published label if any; split key;
execution availability; and hashes of all code and tests.  For SWE-bench Verified, the
manifest additionally records the issue ID, reference and candidate patch
identifiers, test command and outcome, isolated execution harness, and any
generated counterexample.

\subsection{Execution-grounded verification}

A surrogate trace proposes a witness; it is not the verification oracle.
Every reported disagreement is audited against immutable program revisions as
follows:
\begin{enumerate}[leftmargin=*,itemsep=1pt,topsep=2pt]
  \item Check that the proposed input lies in $\domain$ and that the differing
  field is enabled by $\obs$.
  \item Execute both sides under the same recorded environment contract and
  resource policy.  Return values, exceptions, selected mutations, and effects
  are serialized with the comparators in Table~\ref{tab:observable-schema}.
  \item Classify the audit as \emph{confirmed} only when direct observations
  reproduce an in-scope difference; otherwise record \emph{not reproduced} or
  \emph{inconclusive}, together with the failed precondition, unreachable path,
  unresolved symbol, timeout, or harness limitation.
\end{enumerate}

Verification is intentionally asymmetric.  If an EquiBench pair labeled
equivalent yields an executed counterexample, the published label is
demonstrably incorrect for the declared scope.  If a pair labeled inequivalent
does not yield a counterexample, the absence of one does not prove equivalence;
we retain the published label and count the disagreement against \method.  For
SWE-bench Verified, an executed difference establishes non-equivalence to the reference
patch, but does not by itself establish that the candidate violates the issue
intent.  This distinction prevents reference-patch divergence from being
silently promoted to a task-level correctness judgment.

\subsection{Audit trail and failure attribution}

Each challenged pair retains the published label or test status, method
verdict, proposed witness, graph citations, predicted traces, direct
observations, first mismatching observable, verification status, and reason
codes.  Original benchmark labels are never overwritten; any corrected
interpretation is stored as a provenance-preserving amendment.

Unconfirmed and unresolved cases are attributed to the component that blocks
the claim: graph construction or symbol resolution; invalid or out-of-domain
input; infeasible surrogate trace; nonlocal compensation in the full patch;
unsupported environment behavior; execution-harness failure; or an ambiguous
specification.  This record separates benchmark defects from method errors and
from cases that available evidence cannot decide.

\section{Claim Boundaries, Safety, and Reproducibility}

\paragraphhead{What the evidence establishes.}
A concrete, execution-confirmed counterexample establishes inequivalence for
the declared domain, observables, and environment.  In contrast, an
\vequiv verdict is a bounded empirical conclusion based on covered obligations
and generated inputs, not a proof of universal functional equivalence.
Moreover, because unflagged pairs were not exhaustively executed, the reported
audits estimate the precision of challenged cases rather than overall recall.

\paragraphhead{Residual model and representation risk.}
Blinding reduces direct anchoring between the source and target predictions,
but two surrogates can still share the same mistaken assumption.  Likewise,
the code graph can omit or misresolve dynamic behavior.  Typed unresolved
symbols, trace validation, repeated simulation for unstable cases, and
\vabstain outcomes expose some of these failures; none makes surrogate
execution a sound verifier.  Direct execution, symbolic execution, or formal
verification remains preferable when it is available and applicable.

\paragraphhead{Security and privacy.}
The execution audit runs the two revisions in separate subprocesses under a
recorded timeout.  Running untrusted code additionally requires process,
filesystem, memory, and network isolation appropriate to the deployment.
Proprietary-code deployments require an approved model endpoint, local
inference, or source-redaction policy, and evidence records should omit secrets
embedded in literals, configuration, and runtime state.

\paragraphhead{Reproducibility record.}
The run and pair records defined in Appendix~\ref{app:protocol} are sufficient
to identify the exact inputs to graph construction, model inference,
reconciliation, and direct execution.  Reproduction additionally requires the
program snapshots, pair manifests, parser and model versions, provider access
dates, decoding and budget settings, execution environment, raw evidence
ledgers, and scripts that regenerate reported tables and figures.  Parse
failures, exclusions, timeouts, non-reproduced flags, and abstentions remain in
the machine-readable output rather than being dropped from the audit trail.



\end{document}